\documentclass[runningheads]{llncs}
\usepackage[T1]{fontenc}
\usepackage{graphicx,verbatim}
\usepackage{multirow}%
\usepackage{amsmath,amssymb,amsfonts}%
\usepackage{mathrsfs}%
\usepackage[title]{appendix}%
\usepackage{xcolor}%
\usepackage{textcomp}%
\usepackage{manyfoot}%
\usepackage{booktabs}%
\usepackage{algorithm}%
\usepackage{algorithmicx}%
\usepackage{algpseudocode}%
\usepackage{listings}%
\usepackage{adjustbox}
\usepackage{pifont}
\usepackage{array} 
\usepackage{caption}
\usepackage{lineno}
\usepackage{tabularx}
\usepackage{enumitem}
\usepackage{hyperref}
\usepackage{siunitx}
\usepackage{threeparttable}
\usepackage{pgfplots}
\pgfplotsset{compat=1.18}
\usepackage{pgfplotstable}
\usepackage{makecell}

\usepackage{color}

\begin{document}
\title{End2Reg: Learning Task-Specific Segmentation for Markerless Registration in Spine Surgery}
\titlerunning{End2Reg}

\author{Lorenzo Pettinari\inst{1}\orcidID{0009-0009-1722-9024} %index{Pettinari, Lorenzo}
\and
Sidaty El Hadramy\inst{1}\orcidID{0009-0000-2917-0706} %index{El Hadramy, Sidaty}
\and
Michael Wehrli \inst{1}\orcidID{0009-0005-8740-3295} %index{Wehrli, Michael}
\and
Philippe C. Cattin \inst{1}\orcidID{0000-0001-8785-2713} %index{Cattin, Philippe C.}
\and
Daniel Studer \inst{2}\orcidID{0000-0003-1893-4859} %index{Studer, Daniel}
\and
Carol C. Hasler \inst{2} %index{Hasler, Carol C.}
\and
Maria Licci \inst{2}\orcidID{0000-0002-2625-4151}} %index{Licci, Maria}
\authorrunning{Pettinari et al.}

\institute{Department of Biomedical Engineering, University of Basel, Allschwil, Switzerland \email{lorenzo.pettinari@unibas.ch} \and
Department of Orthopedics, University Children’s Hospital, Basel, Switzerland}

\maketitle              % typeset the header of the contribution
\begin{abstract}
Intraoperative navigation in spine surgery demands millimeter-level accuracy. Currently, this is achieved through radiation-intensive intraoperative imaging and bone-anchored markers that are invasive and disrupt surgical workflow. Markerless RGB-D registration methods offer a promising alternative. However, existing approaches rely on weak segmentation labels to isolate relevant anatomical structures, potentially propagating errors through the registration process. We present End2Reg, an end-to-end deep learning framework that jointly optimizes segmentation and registration, eliminating the need for segmentation labels and manual steps. The network learns task-specific segmentation masks optimized for registration, guided solely by the registration objective without explicit segmentation supervision. End2Reg achieves state-of-the-art performance on ex- and in-vivo benchmarks, reducing median Target Registration Error by 32\% and mean Root Mean Square Error by 61\%, while maintaining robust performance under partial occlusions. Ablation results confirm that end-to-end optimization significantly improves registration accuracy. Overall, End2Reg advances towards fully automatic, markerless intraoperative navigation.
Code and interactive visualizations are available at: \href{https://lorenzopettinari.github.io/end-2-reg/}{https://lorenzopettinari.github.io/end-2-reg/}.


\end{abstract}

\section{Introduction}\label{introduction}

Intraoperative navigation systems are essential in orthopedic surgery, where millimeter-level accuracy is crucial for safety and clinical success \cite{luther2015comparison}. This is particularly evident in pedicle screw insertion, where misplacement can cause neural or vascular injury \cite{gelalis2012accuracy,karkenny2019role}. Effective navigation relies on robust registration to align image-derived anatomy with the surgical coordinate system, enabling instrument tracking \cite{floyd2020review,karkenny2019role}.
Preoperative imaging such as CT or MRI allows detailed surgical planning, including implant selection and trajectory optimization \cite{floyd2020review}. For integration into navigation systems, accurate alignment between preoperative images and intraoperative anatomy is essential. Current approaches typically achieve this alignment via intraoperative 3D radiographic imaging and bone-anchored fiducial markers \cite{holly2003intraoperative,karkenny2019role}. While effective, these strategies have several drawbacks: ionizing radiation exposure for patients and staff, particularly when repeated scans are required in multi-level procedures \cite{striano2024intraoperative}; high costs and a large operating room footprint; prolonged workflow \cite{tonetti2020role}; and invasive markers, which can loosen during surgery and require repeated registrations \cite{holly2003intraoperative}.
To overcome these limitations, recent work has investigated radiation-free and markerless approaches, often using RGB-D cameras to capture the exposed surgical surface. RGB-D sensors are compact, low-cost and easy to integrate into the operating room. In these methods, intraoperative video frames are converted into 3D point clouds, while preoperative scans are segmented and sampled to obtain point clouds of the anatomy (e.g., the spine). Registration is then formulated as a point cloud alignment problem between pre- and intraoperative data \cite{daly2025towards,hu2021occlusion,liebmann2024automatic,liu2020automatic}.

\begin{figure}[t]
    \centering
    \includegraphics[width=0.9\textwidth]{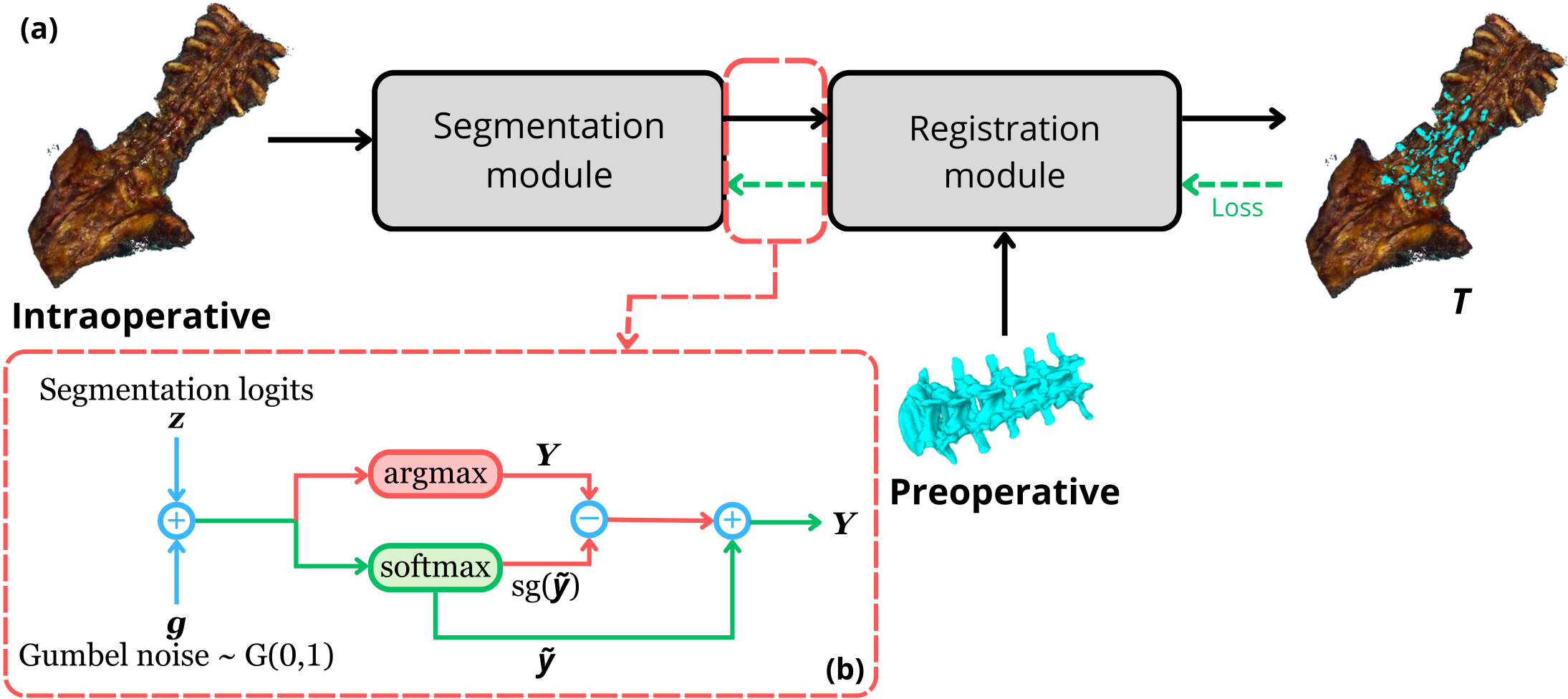}
    \caption{Overview of the \textbf{End2Reg} framework. (a) Network architecture integrates segmentation and registration to estimate the rigid transformation $T$ aligning intraoperative RGB-D-derived point clouds and preoperative point clouds. (b) Straight-Through Gumbel-Softmax (ST-GS) estimator enabling \textbf{end-to-end optimization}, with discrete sampling in the forward pass and soft gradients in the backward pass (Figure inspired by Wei \textit{et al.} \cite{wei2023generalized}).}
    \label{fig:Network_architecture}
\end{figure}

\noindent \textbf{Problem formulation}:
In this context, intraoperative registration can be formulated as a \textbf{heterogeneous point cloud registration} problem performed independently for each RGB-D video frame. Let
$\mathcal{P} = \{\mathbf{p}_i \in \mathbb{R}^3\}_{i=1}^{M}$ be the \textit{preoperative source point cloud}, and
$\mathcal{Q} = \{\mathbf{q}_j \in \mathbb{R}^3\}_{j=1}^{N}$ the \textit{intraoperative target point cloud}.
Our objective is to estimate a rigid transformation
$T = (\mathbf{R}, \mathbf{t}) \in SE(3)$ that aligns $\mathcal{P}$ to $\mathcal{Q}$, thereby minimizing geometric registration errors such as Target Registration Error (TRE) or Root Mean Square Error (RMSE).
This task is challenging due to the \textbf{heterogeneous nature} of the data: preoperative and intraoperative point clouds are extracted from different modalities, and intraoperative clouds can include soft tissue \textbf{partially obscuring bone}, have \textbf{low overlap} due to limited surgical exposure, and be \textbf{corrupted by occlusions} from instruments or the surgeon \cite{daly2025towards,hu2021occlusion,liebmann2024automatic,zhu2020markerless}.

\noindent \textbf{Related work:}
Point cloud registration has been extensively studied \cite{lyu2024rigid}. Traditional methods include ICP \cite{besl1992method}, which is sensitive to initialization and degrades with low overlap, and RANSAC applied to hand-crafted descriptors such as FPFH \cite{fischler1981random,rusu2009fast}, which improves robustness to outliers and partial overlap but depends on reliable feature extraction. Recently, deep learning-based methods have improved robustness under noise and partial overlap \cite{huang2021predator,lyu2024rigid,mei2023overlap,pan2024robust,qin2023geotransformer}, nevertheless, their use on heterogeneous surgical point clouds remains limited \cite{weber2024deep}.
In orthopedic surgery, RGB-D-based registration pipelines typically rely on ICP combined with segmentation to isolate the bony anatomy from the surgically exposed surface \cite{daly2025towards,hu2021occlusion,ji2015patient,liebmann2024automatic,liu2020automatic,WarWil_SparseXM_MICCAI2025,zhu2020markerless}. 
When manual annotations are unavailable, segmentation is commonly supervised using automatically generated weak bone labels obtained from the ground-truth-aligned preoperative anatomy \cite{daly2025towards,hu2021occlusion,liebmann2024automatic,liu2020automatic,WarWil_SparseXM_MICCAI2025}. Although effective for large-scale data generation, these labels are noisy and can propagate errors into the registration pipeline \cite{WarWil_SparseXM_MICCAI2025}.
Additionally, these methods often still require a manual step to select the expected region of overlap on the preoperative anatomy \cite{hu2021occlusion,ji2015patient,liebmann2024automatic,zhu2020markerless}.

\noindent \textbf{Contributions}:
We, therefore, propose \textbf{End2Reg} (\autoref{fig:Network_architecture}), an end-to-end deep learning framework for vertebral registration. Our main contributions are:
\begin{itemize}[leftmargin=*,nosep]
    \item \textbf{Joint learning of segmentation and registration:} segmentation and registration models are trained end-to-end without segmentation labels, allowing End2Reg to learn task-specific segmentations optimized for registration.  
    \item \textbf{State-of-the-art registration performance:} validated on the publicly available SpineDepth \cite{liebmann2021spinedepth} (\textit{ex vivo}) and SpineAlign \cite{daly2025towards} (\textit{in vivo}) datasets.
    \item \textbf{Robustness to occlusions:} End2Reg demonstrates improved robustness under partial occlusions, compared to existing methods.
\end{itemize}

\section{Method}\label{methods}

To address limitations of current RGB-D registration pipelines in orthopedic surgery, we
train segmentation and registration end-to-end. By optimizing segmentation through the registration loss, \textbf{End2Reg} learns task-specific segmentations tailored for downstream registration without any segmentation label. 
An overview of the architecture is shown in \autoref{fig:Network_architecture} and the following subsections describe its main components.

\subsection{Segmentation Module} The segmentation module processes target point clouds \(Q\), in which each point is defined by its 3D coordinates and normalized RGB color values. Its task is to predict a binary mask \(\mathbf{Y}\) identifying regions relevant for registration (e.g., bony anatomy). 
We implement it using KPConv \cite{thomas2019kpconv} in a U-Net-like encoder-decoder architecture. We chose KPConv due to its ability to extract features from irregular and sparse point clouds \cite{thomas2019kpconv}. Given an input point cloud \(Q\) with \(N\) points, the network outputs logits \(\mathbf{z}_j = [z_{j}^{0}, z_{j}^{1}]\) for each point \(j = 1, \ldots, N\), corresponding to the irrelevant and relevant classes, respectively.
Applying a softmax yields class probabilities $\boldsymbol{\pi}_j = \mathrm{softmax}(\mathbf{z}_j)$.
A binary segmentation mask $\mathbf{Y} = [Y_1, \ldots, Y_N]$ is obtained by assigning each point to the class with the highest probability, where $Y_j \in \{0,1\}$ denotes the predicted label of point $j$. This mask is then used as the per-point feature input to the registration module.

\subsection{Registration Module} The registration module takes the source and target point clouds \(P\) and \(Q\) and predicts the rigid transformation \(T\) that aligns them.
We base it upon GeoTransformer as this demonstrated superior registration accuracy compared to other tested methods, is robust to low-overlap, and incorporates outlier filtering \cite{qin2023geotransformer}.
GeoTransformer builds upon KPConv \cite{thomas2019kpconv} for feature extraction. In the standard KPConv formulation, when no additional features are available, each point is initialized with a constant scalar feature equal to one, forcing the network to rely solely on geometric relationships between neighboring points. Following this convention, points in the source cloud \(P\) are assigned a constant feature of 1. For the target point cloud \(Q\), we instead leverage the mask \(\mathbf{Y}\) produced by the segmentation module. Specifically, each point \(\mathbf{q}_j \in Q\) is assigned a binary feature \(F(\mathbf{q}_j) = Y_j \in \{0,1\}\).
The KPConv operation at a query point $\mathbf{q}$ can be expressed as
\begin{equation}
(F * g)(\mathbf{q}) = \sum_{\mathbf{q}_j \in \mathcal{N}_q} g(\mathbf{q}_j - \mathbf{q}) \, F(\mathbf{q}_j),
\label{eq:masked_kpconv}
\end{equation}
where $\mathbf{q}_j$ denotes the position of a neighboring point, $\mathcal{N}_q$ is the neighborhood of $\mathbf{q}$, and $g(\cdot)$ is the KPConv kernel function.
In this formulation, all points of \(Q\) remain part of the geometry, but only those with $Y_j=1$ contribute to the convolution output. This design ensures that the registration module focuses on regions identified by \(\mathbf{Y}\) as relevant.

\subsection{End-to-End Training}
In our formulation, the segmentation module outputs a per-point binary mask $\mathbf{Y}$ that conditions the registration module. Hard mask generation via $\arg\max$ yields a deterministic, non-differentiable assignment. As a result, \textbf{gradients from the registration loss cannot propagate back to the segmentation module}. 
To enable end-to-end training, we adopt the \textbf{Gumbel-Softmax estimator} \cite{jang2016categorical}, which provides a differentiable approximation of categorical sampling via the Gumbel-Max trick. Specifically, by adding i.i.d. noise $g_{j}^{i} \sim \operatorname{}{Gumbel}(0,1)$ to each logit $z_{j}^{i}$, we obtain $Y_j = \arg\max_i (z_{j}^{i} + g_{j}^{i})$, which is a single sample drawn from the categorical distribution defined by $\boldsymbol{\pi}_j$. 
We then introduce a differentiable relaxation of the categorical sample using the Gumbel-Softmax distribution,
\begin{equation}
\tilde{y}_{j}^{i} = \frac{\exp((z_{j}^{i} + g_{j}^{i})/\tau)}{\sum_{c=0}^{1} \exp((z_{j}^{c} + g_{j}^{c})/\tau)}, \quad i \in \{0,1\},\quad  j = 1,...,N
\end{equation}
where $\tau > 0$. Unlike $\mathbf{Y}$, \textbf{$\mathbf{\tilde{y}}$ is continuous and differentiable with respect to $\mathbf{z}$}. As $\tau \to 0$, $\mathbf{\tilde{y}}$ converges in distribution to a one-hot vector, recovering categorical samples; for larger $\tau$, it produces smoother values. 
To combine the discrete behavior of $\mathbf{Y}$ with the differentiability of $\mathbf{\tilde{y}}$, we employ the \textbf{Straight-Through Gumbel-Softmax Estimator (ST-GS)} \cite{bengio2013estimating,jang2016categorical}, as illustrated in \autoref{fig:Network_architecture}, following the standard \textbf{Straight-Through Estimator} \cite{bengio2013estimating} formulation:

\begin{equation}
\label{eq:ste}
\mathbf{Y_{\mathrm{STE}}} = \mathbf{Y} - \mathrm{sg}(\mathbf{\tilde{y}}) + \mathbf{\tilde{y}}.
\end{equation}
where $\mathrm{sg}(\cdot)$ denotes the stop-gradient operator, which treats its input as a constant during backpropagation. In the \textbf{forward pass}, $\mathbf{Y_{\mathrm{STE}}} = \mathbf{Y}$, because $\mathbf{\tilde{y}} - \mathrm{sg}(\mathbf{\tilde{y}}) = 0$. During \textbf{backpropagation}, the gradient of $\mathbf{Y_{\mathrm{STE}}}$ w.r.t. $\mathbf{z}$ is approximated as:
\begin{equation}
\frac{\partial \mathbf{Y_{\mathrm{STE}}}}{\partial \mathbf{z}} 
= \frac{\partial (\mathbf{Y} - \mathrm{sg}(\mathbf{\tilde{y}}) + \mathbf{\tilde{y}})}{\partial \mathbf{z}}
= \underbrace{\frac{\partial \mathbf{Y}}{\partial \mathbf{z}}}_{=0} 
- \underbrace{\frac{\partial \mathrm{sg}(\mathbf{\tilde{y}})}{\partial \mathbf{z}}}_{=0} 
+ \frac{\partial \mathbf{\tilde{y}}}{\partial \mathbf{z}}
\approx \frac{\partial \mathbf{\tilde{y}}}{\partial \mathbf{z}}.
\end{equation}

The end-to-end framework is trained with the dual-phase coarse-to-fine loss of GeoTransformer \cite{qin2023geotransformer}. Further details can be found in Qin \textit{et al.} \cite{qin2023geotransformer}.

\section{Experiments and Results}\label{Experiments}

\subsection{Datasets and implementation details}

\textbf{End2Reg} was evaluated on two benchmark datasets for vertebral registration: SpineDepth \cite{liebmann2021spinedepth} (\textit{ex vivo}) and SpineAlign \cite{daly2025towards} (\textit{in vivo} and limited surgical exposure), covering both controlled and realistic settings. In the following, we provide details about these datasets as well as their preprocessing:

\begin{itemize}
    \item \textbf{SpineDepth (\textit{ex vivo})} \cite{liebmann2021spinedepth} consists of RGB-D recordings of pedicle screw placement on ten cadaver specimens, with preoperative 3D vertebral models extracted from CT. Ground-truth vertebral poses relative to the RGB-D sensor are provided for each frame via optical tracking. In contrast to Liebmann \textit{et al.} \cite{liebmann2024automatic}, we retain frames with up to $50\%$ occlusion and evaluate the full dataset without viewpoint filtering to reflect realistic intraoperative scenarios. An expert surgeon manually annotated anatomical landmarks (spinous, left and right transverse processes) for each vertebra following Liebmann \textit{et al.} \cite{liebmann2024automatic}; these annotations were used to compute the TRE. Evaluation was performed using 8-fold cross-validation.

    \item \textbf{SpineAlign (\textit{in vivo})} \cite{daly2025towards} contains 24 lumbar open spine surgeries, with preoperative 3D vertebral models extracted from CT or MRI scans.
    Coarse alignment between pre- and intraoperative point clouds is provided using anatomical landmarks identified by the surgeon. Following Daly \textit{et al.} \cite{daly2025towards}, $20$ patients were used for training and $4$ for testing, using the same data split.
\end{itemize}

In both datasets, point clouds were normalized to a unit sphere, and random rigid transformations were applied to each frame to simulate anatomical misalignment. Similar to Pan \textit{et al.} \cite{pan2024robust} and Qin \textit{et al.} \cite{qin2023geotransformer}, translations were limited to 10\% of the point cloud size (\(\approx\)\SI{51}{\mm} on SpineDepth, \(\approx\)\SI{13}{\mm} on SpineAlign) and rotations to a maximum of 45°, as larger misalignments are usually corrected by generic initialization (e.g., aligning center-of-mass or PCA). The training was performed on an NVIDIA RTX 5090 GPU (32\,GB) with a batch size of 1, an initial learning rate of $10^{-4}$, a \num{10000}-iteration warm-up, and cosine annealing decay.

\subsection{Results}

\begin{figure}[htb]
    \centering
    \includegraphics[width=\textwidth]{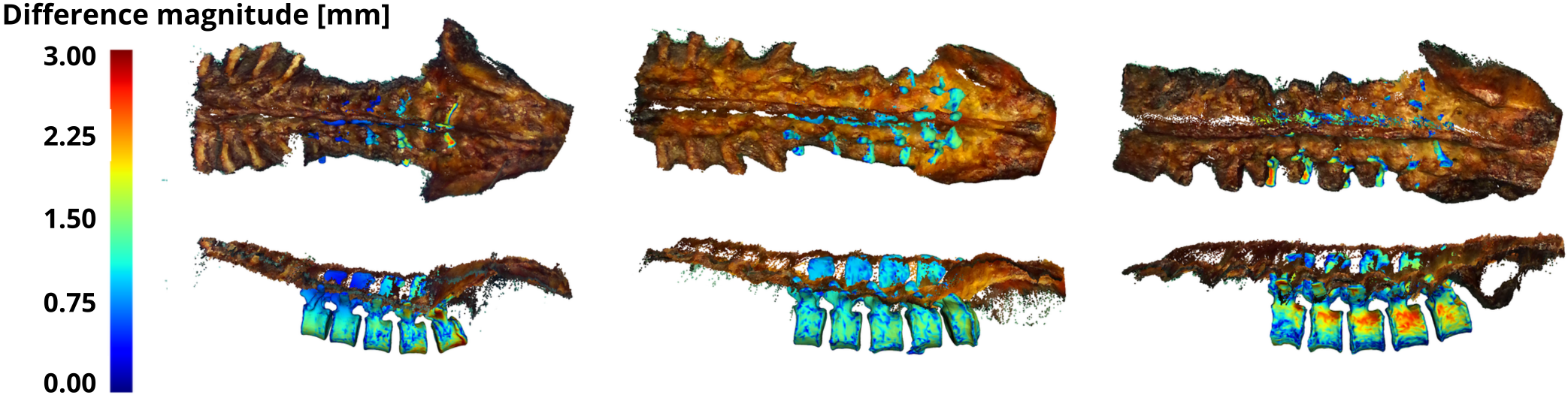}
    \caption{\textbf{End2Reg registration results} of the preoperative to the intraoperative point cloud on three SpineDepth specimens. Colors indicate the per-point displacement magnitude relative to the ground-truth pose.}
    \label{fig:registration_results}
\end{figure}

\begin{figure}[htb]
    \centering
    \includegraphics[width=0.52\textwidth]{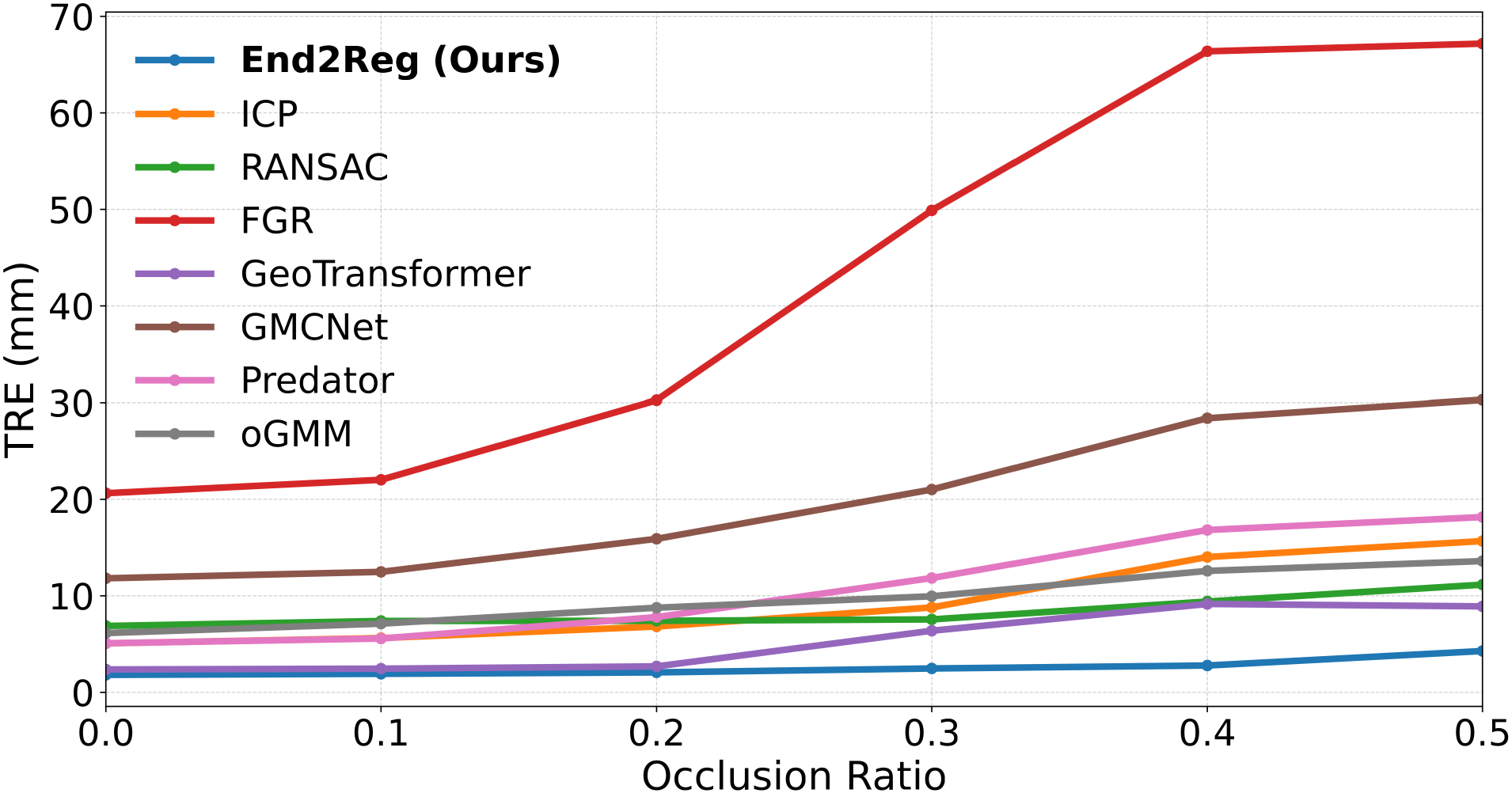}
    \caption{\textbf{TRE under increasing occlusion.} Comparison of different methods as the occlusion ratio increases from 0 to 0.5. End2Reg demonstrates the highest robustness, showing the smallest TRE across all occlusion levels.}
    \label{fig:occlusion_analysis}
\end{figure}

\noindent \textbf{SpineDepth registration results}: \autoref{tab:comparison_spine_combined} presents the TRE along with translation (RTE) and rotation (RRE) errors for our method, compared to the state-of-the-art results reported by Liebmann \textit{et al.} (AutoReg) \cite{liebmann2024automatic} using the same dataset split. The qualitative results are illustrated in \autoref{fig:registration_results}. In addition, to further prove the robustness of our approach to occlusion, we report the TRE as a function of the occlusion ratio in \autoref{fig:occlusion_analysis}.

\noindent \textbf{SpineAlign registration results}: \autoref{tab:comparison_spine_combined} reports the registration results, compared to the state-of-the-art reported by Daly \textit{et al.} (CorrNet) \cite{daly2025towards}. The dataset provides only a coarse pre-to-intraoperative alignment, which Daly \textit{et al.} use as initialization, whereas our approach operates without relying on this prior. As no reliable ground truth is available, the results should be interpreted as a preliminary benchmark. Following Daly \textit{et al.}~\cite{daly2025towards}, we report RMSE and fitness.

\begin{table}[htb]
\centering
\caption{Registration results. SpineDepth: RRE (deg), RTE (mm), TRE (mm) reported as median [Q1, Q3], following Liebmann \textit{et al.} \cite{liebmann2024automatic}. SpineAlign: Fitness, RMSE (mm) reported as mean (± st. dev.), following Daly \textit{et al.} \cite{daly2025towards}.}
\label{tab:comparison_spine_combined}

\begin{threeparttable}
\begin{adjustbox}{max width=\textwidth, center}
\begin{tabular}{l | c c c c | c c c}
\toprule
\textbf{Method}
& \multicolumn{4}{c}{\textbf{SpineDepth}}
& \multicolumn{3}{c}{\textbf{SpineAlign}} \\
\cmidrule(lr){2-5} \cmidrule(lr){6-8}
& \textbf{RRE (deg)}
& \textbf{RTE (mm)}
& \textbf{TRE (mm)}
& \textbf{t (ms)}
& \textbf{Fitness}
& \textbf{RMSE (mm)}
& \textbf{t (ms)} \\
\midrule

\multicolumn{6}{l}{\textit{Classical methods}} \\

ICP${(1)}$
& 2.3[1.5, 3.4]
& 5.0[3.4, 6.7]
& 5.4[3.8, 7.2]
& 339
& .49(±.25)
& 12.26(±4.51)
& 224\\

RANSAC+ICP${(1)}$
& 1.9[1.2, 4.0]
& 3.7[2.3, 5.7]
& 4.2[2.7, 6.7]
& 490
& .51(±.27)
& 12.70(±6.42)
& 365 \\

FGR+ICP${(1)}$
& 2.3[1.3, 11.5]
& 15.9[10.3, 30.6]
& 17.2[10.8, 37.2]
& 430
& .21(±.18)
& 17.89(±5.13)
& 286\\

\addlinespace
\multicolumn{6}{l}{\textit{Deep learning-based methods}} \\

GMCNet \cite{pan2024robust}
& 11.1[7.1, 17.1]
& 10.0[7.2, 13.9]
& 12.7[9.5, 17.1]
& 400
& .22(±.20)
& 9.31(±2.91)
& 280\\

GeoTransformer \cite{qin2023geotransformer}
& 1.4[1.0, 2.0]
& 1.9[1.2, 3.1]
& 2.2[1.5, 3.4]
& 638
& .70(±.12)
& 3.04(±1.76)
& 455 \\

OverlapPredator \cite{huang2021predator}
& 2.8[1.9, 4.1]
& 4.8[3.5, 6.6]
& 5.1[3.8, 7.0]
& 960
& .34(±.18)
& 9.93(±4.58)
& 485 \\

oGMM \cite{mei2023overlap}
& 3.1[1.8, 5.5]
& 6.4[4.0, 9.2]
& 6.7[4.3, 9.7]
& 1150
& .14(±.09)
& 9.73(±2.48)
& 765 \\

AutoReg \cite{liebmann2024automatic}${(1,2)}$
& --
& --
& 2.7[1.7, 3.6]
& --
& --
& --
& -- \\

CorrNet \cite{daly2025towards}${(3)}$
& --
& --
& --
& --
& .58(±.11)
& 7.14(±0.47)
& -- \\

\midrule
\textbf{End2Reg (Ours)}
& \textbf{1.3[0.8, 1.8]}
& \textbf{1.9[1.2, 2.7]}
& \textbf{1.8[1.2, 2.7]}
& 620
& \textbf{.71(±.18)}
& \textbf{2.80(±1.31)}
& 450 \\

\bottomrule
\end{tabular}
\end{adjustbox}
\\[1mm] 
\raggedright
(1) Requires manual selection of expected overlapping regions on SpineDepth data.\\
(2) Results obtained on a filtered dataset with specific viewpoints and no occlusions.\\
(3) Requires manual initialization of the registration.

\end{threeparttable}
\end{table}

\noindent \textbf{Further comparing methods}:
We additionally compared \textbf{End2Reg} against a set of segmentation-registration baselines. All the baselines use the same KPConv-based segmentation network as End2Reg while differing in the registration module. In contrast to End2Reg, segmentation and registration networks were trained independently.
The segmentation module was trained using a cross-entropy loss with weak bone labels, generated by assigning points in \(Q\) within a threshold distance of the surface \(P\) aligned to the ground-truth pose \cite{daly2025towards,WarWil_SparseXM_MICCAI2025}. Thresholds of \SI{3}{\milli\metre} for SpineDepth and \SI{10}{\milli\metre} for SpineAlign were used, yielding segmentation performance (\autoref{tab:seg_spinedepth}) consistent with prior work \cite{daly2025towards,liebmann2024automatic}.
For registration, we evaluated both classical \cite{besl1992method,fischler1981random,zhou2016fast} and learning-based methods \cite{huang2021predator,mei2023overlap,pan2024robust,qin2023geotransformer}. Consistent with our problem formulation, we selected learning-based rigid registration methods designed for low-overlap scenarios. To aid classical baselines on SpineDepth, we follow prior work \cite{liebmann2024automatic} and include an additional manual step consisting of selecting the expected region of overlap on the preoperative anatomy.

\begin{figure}[htb]
    \centering
    \includegraphics[width=0.9\textwidth]{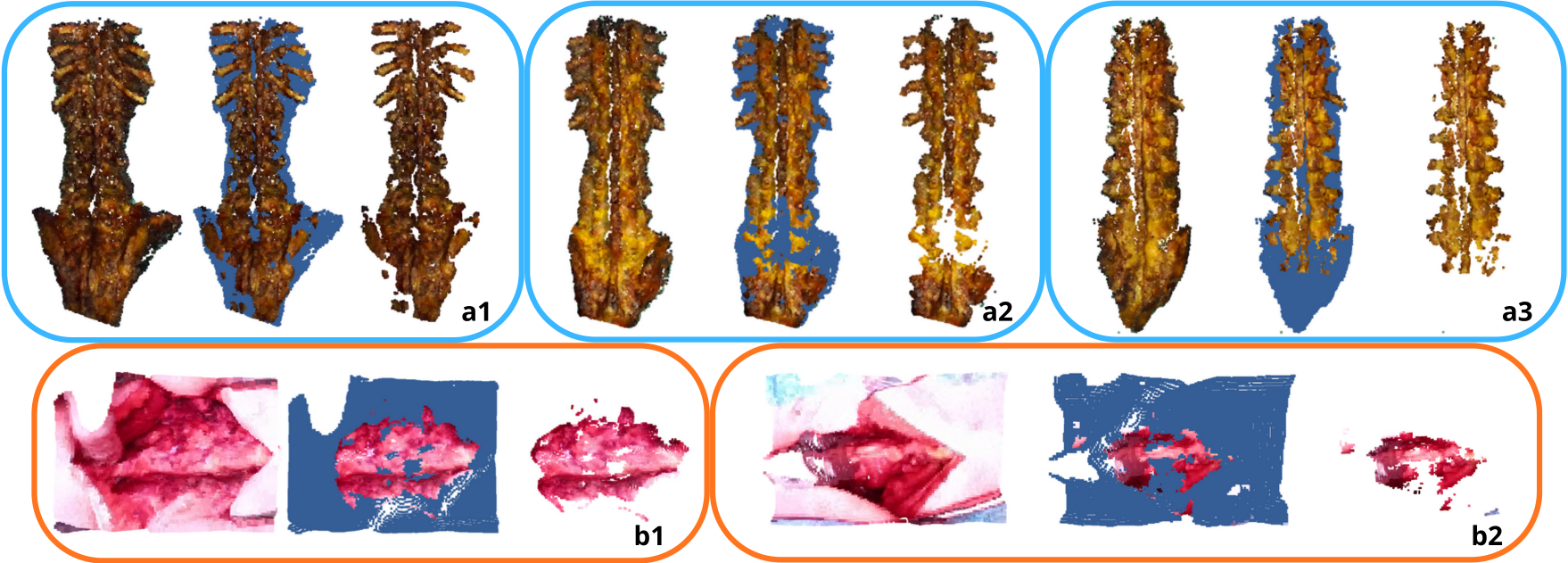}
    \caption{Task-specific segmentations on three specimens for SpineDepth (a) and two cases on SpineAlign (b). For each, three point clouds are shown: intraoperative input point cloud, segmentation (background in blue), and registration-relevant points, which mainly correspond to bone structures.}
    \label{fig:Unsupervised_segmentation}
\end{figure}

\noindent \textbf{Task-specific segmentation}: Qualitative results (\autoref{fig:Unsupervised_segmentation}) show that our model highlights regions informative for registration, corresponding primarily to bony anatomy while suppressing surrounding soft tissue. Importantly, these task-specific segmentations are not intended to reproduce anatomical ground truth or weak bone labels but are optimized for downstream registration. Accordingly, overlap-based metrics are not used as primary evaluation criteria. Instead (\autoref{tab:seg_spinedepth}), we report the one-sided Chamfer distance and the 95th percentile Hausdorff distance from the weak-label point cloud to the predicted segmentation, measuring the containment of weak bone labels.

\begin{table}[thb]
\centering
\captionsetup{width=\textwidth}
\caption{Task-specific segmentation quantitative evaluation: Dice, IoU, one-sided Chamfer (mm) and HD95 (mm). 
Results as median [Q1, Q3] for SpineDepth, as Liebmann \textit{et al.} \cite{liebmann2024automatic}, and mean (± st. dev.) for SpineAlign, as Daly \textit{et al.} \cite{daly2025towards}.}
\label{tab:seg_spinedepth}
\begin{adjustbox}{max width=\textwidth, center}
\begin{tabular}{l | ccc | ccc}
\toprule
\textbf{Method} 
& \multicolumn{3}{c}{\textbf{SpineDepth}} 
& \multicolumn{3}{c}{\textbf{SpineAlign}} \\
\cmidrule(lr){2-4} \cmidrule(lr){5-7}
& \textbf{Dice}$\uparrow$
& \textbf{Chamfer}$\downarrow$
& \textbf{HD95}$\downarrow$
& \textbf{IoU}$\uparrow$
& \textbf{Chamfer}$\downarrow$
& \textbf{HD95}$\downarrow$ \\
\midrule
AutoReg \cite{liebmann2024automatic}
& \textbf{.74[.72, .74]} & - & -
& - & - & - \\
CorrNet \cite{daly2025towards}
& - & - & -
& .78(±.00) & - & - \\
KPConv 
& .73[.68, .77] & 1.5[1.0, 2.4] & 10.5[7.2, 15.3]
& \textbf{.84(±.05)} & \textbf{0.5(±1.3)} & \textbf{3.3(±3.2)} \\
\textbf{End2Reg} 
& .50[.42, .54] & \textbf{0.9[0.6, 1.3]} & \textbf{5.6[4.3, 7.3]}
& .70(±.06) & 0.8(±0.3) & 4.2(±0.9) \\
\bottomrule
\end{tabular}
\end{adjustbox}
\end{table}

\noindent \textbf{Ablation Study}:
We evaluated the effect of end-to-end training by comparing \textbf{End2Reg} with the same architecture trained in a two-step manner, where segmentation and registration are optimized sequentially rather than jointly (reported as \textit{GeoTransformer} in \autoref{tab:comparison_spine_combined}). 
Wilcoxon signed-rank tests on SpineDepth results show that the improvement of End2Reg is significant ($p \ll 0.05$) with a moderate effect size ($r = 0.47$) and reduced outlier ratio (3.8\% vs 7.0\%).

\section{Conclusion}\label{conclusion}

We presented \textbf{End2Reg}, an \textbf{end-to-end framework} for pre-to-intraoperative vertebral point cloud registration achieving state-of-the-art results on SpineDepth (\textit{ex vivo}) \cite{liebmann2021spinedepth} and SpineAlign (\textit{in vivo}) \cite{daly2025towards}. Our method \textbf{removes the need for segmentation labels} by learning task-specific segmentations optimized for registration. It enables accurate alignment of the full preoperative anatomy while remaining robust to occlusions.
Evaluation was limited by available datasets: SpineDepth lacks preoperative deformation, and SpineAlign provides only coarse manual alignment. 
Future work will address intervertebral deformation and acquire an \textit{in vivo} dataset with CT-based ground-truth alignment. Overall, End2Reg shows strong potential to improve accuracy and automation of markerless intraoperative registration, supporting its integration into clinical workflows.

% \begin{credits}
% \subsubsection{\discintname}
% The authors have no competing interests to declare relevant to this article's content.
% \end{credits}

%
% ---- Bibliography ----
%

\bibliographystyle{splncs04}

\bibliography{sn-bibliography}

\end{document}